%% file: main.tex
\documentclass[10pt,a4paper]{article}
\usepackage{wampp-preprint}
\usepackage[hyphens]{url}
\usepackage{graphicx}
\usepackage{natbib}
\usepackage{caption}
\usepackage{algorithm}
\usepackage{algorithmic}
\usepackage{amsmath}
\usepackage{amssymb}
\usepackage{booktabs}
\usepackage{xcolor}
\usepackage{multirow}
\usepackage{makecell}
\usepackage{pifont}
\usepackage{amsfonts}
\usepackage{bm}
\usepackage{colortbl}
\usepackage{amsthm}
\usepackage{subcaption}
\usepackage{mathrsfs}
\usepackage{enumitem}
\usepackage{bibunits}
\graphicspath{{figures/}}
\newcommand{\name}{Latent-WAM++}

\newcommand{\method}{Latent-WAM++}
\newcommand{\sg}{\operatorname{sg}}
\newcommand{\MSE}{\operatorname{MSE}}

\newcommand{\tbd}{\textnormal{--}}
\newcommand{\yesmark}{\ensuremath{\checkmark}}
\newcommand{\nomark}{\ensuremath{\times}}

\title{Latent-WAM++: Shaping World and Action Representations for End-to-End Autonomous Driving}

\author{
Yichen Zhang\textsuperscript{\rm 1,2}\footnotemark[1]\footnotemark[4],
Yupeng Zheng\textsuperscript{\rm 1}\footnotemark[1],
Linbo Wang\textsuperscript{\rm 1},
Pengxuan Yang\textsuperscript{\rm 1},
Zebin Xing\textsuperscript{\rm 1},\\
Junli Wang\textsuperscript{\rm 1},
Shiwei Li\textsuperscript{\rm 2}\footnotemark[2],
Kangwei Liu\textsuperscript{\rm 2}\footnotemark[2],
Fan Ren\textsuperscript{\rm 2},\\
Qichao Zhang\textsuperscript{\rm 1}\footnotemark[3],
Yifeng Pan\textsuperscript{\rm 2}\footnotemark[2],
Ji Tao\textsuperscript{\rm 2},
Dongbin Zhao\textsuperscript{\rm 1}
}
\affiliations{
\textsuperscript{\rm 1}Institute of Automation, Chinese Academy of Sciences\\
\textsuperscript{\rm 2}Chongqing Changan Technology Co., Ltd.
}

\begin{document}

\begin{bibunit}[wampp-natbib]
\setlength{\dbltextfloatsep}{14pt plus 2pt minus 3pt}
\setlength{\textfloatsep}{14pt plus 2pt minus 3pt}
\maketitle
\begingroup
\renewcommand{\thefootnote}{\fnsymbol{footnote}}
\footnotetext[1]{Equal contribution.}
\footnotetext[2]{Project leaders.}
\footnotetext[3]{Corresponding author.}
\footnotetext[4]{This work was done during an internship at Chongqing Changan Technology Co., Ltd.}
\endgroup

\begin{abstract}
\input{src/00_abstract}
\end{abstract}

\input{src/01_introduction}
\input{src/02_related_work}
\input{src/03_method}
\input{src/04_experiments}
\input{src/05_conclusion}

\clearpage
\putbib[main]
\end{bibunit}

\clearpage
\onecolumn
\raggedbottom
\appendix
\setcounter{table}{0}
\setcounter{figure}{0}
\setcounter{equation}{0}
\renewcommand{\thetable}{S\arabic{table}}
\renewcommand{\thefigure}{S\arabic{figure}}
\renewcommand{\theequation}{S\arabic{equation}}
\begin{bibunit}[plainnat]
\input{appendix}
\end{bibunit}

\end{document}

%% file: src/00_abstract.tex
We introduce \method{}, a representation-centered World Action Model (WAM) for end-to-end autonomous driving. Existing WAMs are primarily trained to predict future states or final actions, leaving the intermediate world and action representations that connect prediction to planning only indirectly constrained. \method{} matches the supervision of each representation to its role in planning. World Representation Construction combines scene-ego future prediction, SIGReg-based distribution stabilization, and frozen geometric priors to learn a temporally predictive, non-degenerate, and geometry-aware world representation without an EMA target encoder. Trajectory MeanFlow Planner directly predicts multimodal trajectory proposals while being optimized with an Improved MeanFlow velocity-space objective, enabling one-step sampling. Visual-Action Representation Alignment further aligns the resulting pre-output action features with a stop-gradient target derived from current visual semantics. \method{} achieves state-of-the-art performance on NAVSIM v2 and under zero-shot evaluation on HUGSIM, reaching 90.1 EPDMS on NAVSIM v2 and 47.0 average RC and 30.1 average HD-Score on HUGSIM. Training the planner end to end costs 190 GPU-hours for \method{} versus 1,600 for Latent-WAM, a reduction of 88.1\%.

%% file: src/01_introduction.tex
\section{Introduction}
\label{sec:introduction}

\input{figures/teaser_layout}

Driving World Action Models (WAMs) couple world modeling with action generation to enable end-to-end driving planning~\citep{law,world4drive,epona,dreamerad,latentwam}.
From a representation perspective, a WAM first encodes the observed scene and its anticipated dynamics into a world representation, transforms this representation into an action representation, and subsequently decodes it into future ego trajectories.
The world representation determines what scene information is available for planning, whereas the action representation governs how that information is translated into future trajectories.
Planning performance therefore depends critically on both representations and, more importantly, on how effectively they interact.
This leads to a central question: \textbf{How should world and action representations be jointly shaped to better support WAM-based driving planning?}

Early studies primarily supervise predicted future states or final actions~\citep{law}, leaving the internal representations to emerge as a by-product of these output-oriented objectives.
More recent approaches, such as World4Drive and Latent-WAM, strengthen representation learning through visual foundation models and predictive latent world modeling, respectively~\citep{world4drive,latentwam}.
Nevertheless, existing conditioning exposes world features to the decoder, but does not impose a representation-level alignment on its pre-output action representations.
Such isolated optimization can limit representation learning efficiency and lead to suboptimal planning performance, as shown in Figure~\ref{fig:teaser}.
Specifically, future latent prediction alone cannot prevent the learned feature distribution from collapsing toward a low-variance solution when the encoder and predictor are jointly optimized~\citep{balestriero2025lejepa}.
Moreover, conditioning an action decoder on world representations merely makes scene information accessible; it does not explicitly align the two representation spaces or ensure that the action representation preserves the planning-relevant information encoded by the world model.

These limitations are not isolated implementation details.
Together, they motivate a unified representation-learning problem: learning informative, non-collapsed world representations while explicitly aligning them with action representations, so that future trajectories can fully benefit from knowledge acquired through world modeling.

We address the question through a unified principle: \emph{match the supervision of each intermediate representation to its role in planning}.
The world representation should encode scene dynamics, maintain a non-degenerate distribution across driving scenes, and preserve geometric context.
The action representation should directly parameterize the trajectories ultimately required by the planner.
Their correspondence should be explicitly encouraged without forcing the two representations to share identical semantics.
Guided by this principle, we introduce \textbf{\method{}}, which jointly shapes world and action representations through three coordinated components.
\textbf{World Representation Construction (WoRC)} combines Scene-Ego future prediction with SIGReg~\citep{balestriero2025lejepa}, using direct distribution regularization in place of an EMA-updated target encoder, and incorporates frozen geometric priors.
\textbf{Trajectory MeanFlow Planner (TrajMF)} directly predicts multimodal trajectory proposals and induces an Improved MeanFlow velocity-space objective from these predictions~\citep{improvedmeanflow}, yielding trajectory-predictive action features and enabling one-step generation.
\textbf{Visual-Action Representation Alignment (VARA)} asymmetrically aligns these action features with a stop-gradient target derived from the current visual representation.
Thus, WoRC constructs the world representation, TrajMF defines the action representation, and VARA explicitly regularizes the visual-action interface.

\method{} establishes new state-of-the-art results with 90.1 EPDMS on NAVSIM v2~\citep{navsimv2metrics} and 47.0/30.1 average RC/HD-Score under zero-shot evaluation on HUGSIM~\citep{hugsim}.
As shown in Figure~\ref{fig:teaser}, under the same H20 hardware and data-loading pipeline, training our planner end to end costs $190$ GPU-hours versus $1{,}600$ for Latent-WAM, a reduction of $88.1\%$.

Our contributions are summarized as follows:
\begin{itemize}
    \item We formulate joint representation shaping in WAM-based planning as the combination of role-matched representation objectives and explicit regularization of the visual-action interface.
    
    \item We introduce \method{}, which constructs a predictive, non-degenerate, and geometry-aware world representation, produces trajectory-predictive action features through one-step multimodal generation, and asymmetrically aligns these features with visual semantics.
    
    \item \method{} achieves state-of-the-art results on NAVSIM v2 and zero-shot HUGSIM while reducing the cost of training the planner end to end by $88.1\%$ compared with Latent-WAM ($190$ versus $1{,}600$ GPU-hours).
\end{itemize}

%% file: figures/teaser_layout.tex
\begin{figure}[t]
    \centering
    \includegraphics[width=0.87\columnwidth]{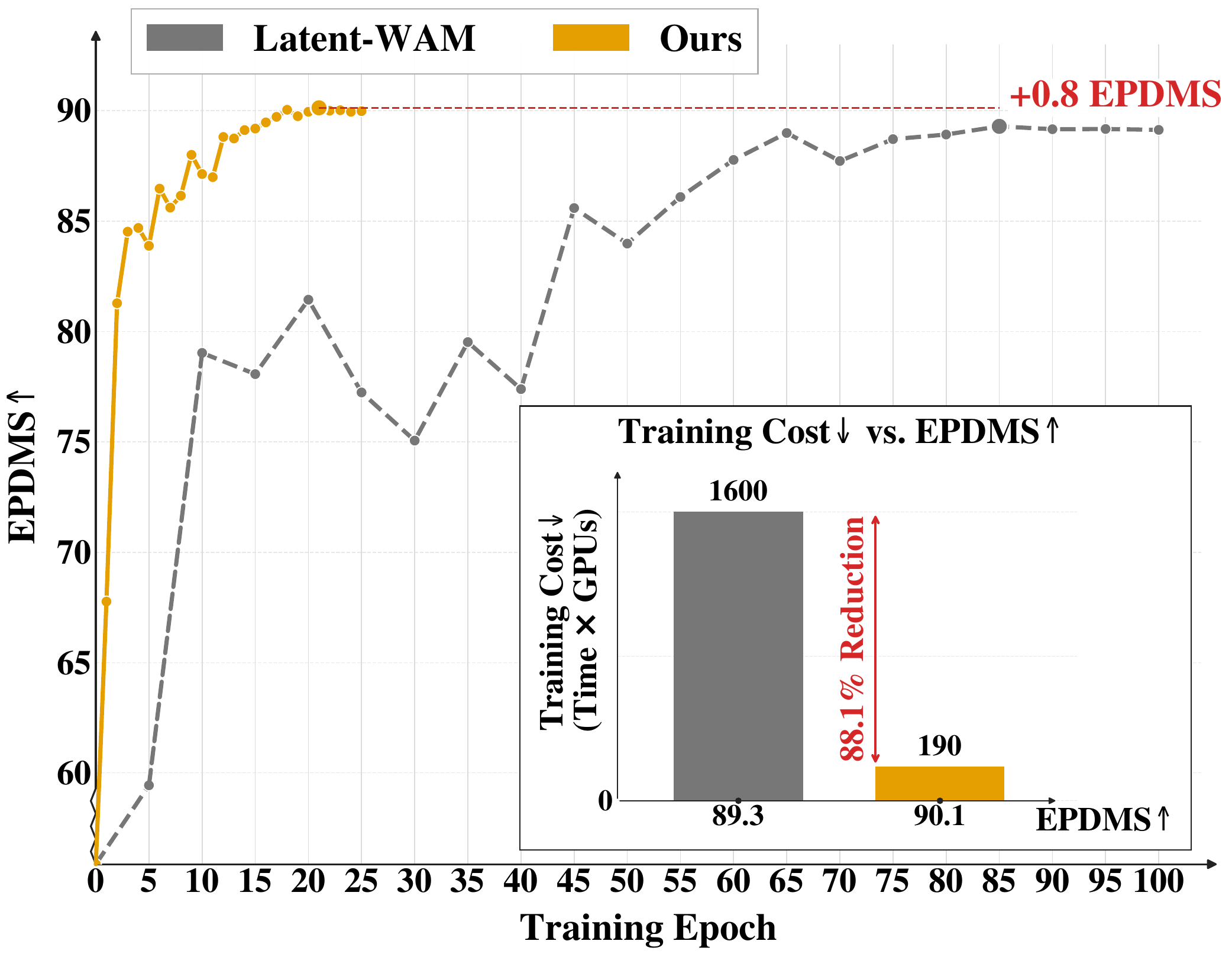}
    \caption{NAVSIM v2 performance and the cost of training planners end to end under identical hardware and data-loading settings. Our planner improves EPDMS by $0.8$ over Latent-WAM ($90.1$ vs.\ $89.3$) and costs $88.1\%$ less to train end to end ($190$ vs.\ $1{,}600$ GPU-hours).}
    \label{fig:teaser}
\end{figure}

%% file: src/02_related_work.tex
\section{Related Work}
\label{sec:related_work}

\subsection{World Models for Autonomous Driving}

Research on driving world models has focused on three properties of planning-oriented representations: capturing temporal evolution, maintaining stable predictive features, and encoding scene geometry.
Some approaches~\citep{drivedreamer,epona,drivelaw} model future scene evolution through video generation, but pixel-level generation incurs substantial computational cost.
Latent world models~\citep{law,world4drive,latentwam} instead predict future states in representation space, providing temporal supervision without reconstructing future images.

Jointly optimizing the encoder and latent predictor without an anti-collapse mechanism may produce degenerate representations~\citep{balestriero2025lejepa}.
Drive-JEPA~\citep{drivejepa} and Latent-WAM~\citep{latentwam} use an EMA-updated, stop-gradient target encoder to prevent representation collapse.
LeWorldModel~\citep{leworldmodel} instead applies SIGReg to stabilize latent learning without maintaining an additional target encoder.
Geometry is incorporated through different forms of supervision and representation.
Latent-WAM~\citep{latentwam} transfers geometric priors into visual features through geometry distillation, while occupancy-based methods~\citep{occworld,sparseworld} represent the driving scene explicitly in structured 3D space.
The former shapes the learned visual features, whereas the latter provides explicit spatial context for planning.
Accordingly, within our latent future-prediction framework, WoRC addresses both representation stability and geometric structure: SIGReg stabilizes latent learning without a target encoder, while occupancy context augments geometry-distilled visual features.

\subsection{Action Decoding for Driving Planning}

Driving planners mainly decode future ego motion through query-based trajectory regression, autoregressive prediction, or generative modeling.
Query-based planners~\citep{law,world4drive,latentwam} directly decode action queries into future trajectories, whereas autoregressive planners~\citep{drivinggpt} sequentially predict discrete action tokens.
Generative planners provide a natural mechanism for modeling multiple plausible futures.
DiffusionDrive~\citep{diffusiondrive} and GoalFlow~\citep{goalflow} generate multimodal trajectories using diffusion and flow-based formulations, respectively.
MeanFuser~\citep{meanfuser} further adapts MeanFlow~\citep{meanflow} with Gaussian-mixture noise for one-step multimodal trajectory generation.
Our TrajMF follows this one-step generative direction and directly parameterizes clean trajectory proposals under Improved MeanFlow optimization~\citep{improvedmeanflow}.

%% file: src/03_method.tex
\section{Method}
\label{sec:method}

\input{figures/overview_layout}

\subsection{Overview}
\label{sec:overview}

Figure~\ref{fig:overview} shows the three components of our method.
World Representation Construction (WoRC) constructs a temporally
predictive and distribution-regularized world representation with
geometric context (Section~\ref{sec:world_representation}). The
Trajectory MeanFlow Planner (TrajMF) retrieves a planning condition and
directly predicts and aggregates multimodal trajectories in one step
(Section~\ref{sec:trajmf}). Its world-conditioned pre-output features form
the action representation, which Visual-Action Representation Alignment
(VARA) relates to current visual semantics
(Section~\ref{sec:vara}).

\subsection{World Representation Construction}
\label{sec:world_representation}

\paragraph{Scene encoding.}

We first convert multi-view image observations into compact visual tokens that
serve as the semantic backbone of the world representation.
Given patch tokens $X\in\mathbb{R}^{T\times V\times S\times D_e}$, we
broadcast $N_q$ learnable queries to each view and frame, obtaining
$Q_v\in\mathbb{R}^{T\times V\times N_q\times D_e}$. A DINOv2 encoder
$\mathcal{E}_v$~\citep{dinov2} and projector $\mathcal{P}_v$ produce
\begin{equation}
[F_v,\widehat{X}]
=
\mathcal{P}_v\!\left(\mathcal{E}_v([Q_v;X])\right),
\label{eq:scene_encoding}
\end{equation}
where $F_v\in\mathbb{R}^{T\times N_v\times D}$, $N_v=VN_q$, and
$\widehat{X}$ denotes projected patch features.

However, DINOv2 features alone may lack the metric spatial cues needed for driving scenes. We therefore enrich the representation with two frozen geometry-aware branches. First, a frozen teacher distills geometric knowledge into the visual tokens without adding trainable geometry modules.
A frozen WorldMirror-based teacher
$\mathcal{E}_{\mathrm{geo}}$~\citep{worldmirror} provides stop-gradient
targets for patch-level distillation:
\begin{equation}
\mathcal{L}_{\mathrm{geo}}
=
1-\operatorname{cos}\!\left(
\operatorname{LN}(\mathcal{P}_g(\widehat{X})),
\sg\!\left(
\operatorname{LN}(\mathcal{E}_{\mathrm{geo}}(I))
\right)
\right),
\label{eq:geo_distillation}
\end{equation}
where $\mathcal{P}_g$ aligns visual patch features with the teacher feature
space.

We then add ego and occupancy cues so that the planner receives both intent
and metric structure at the current frame.
The ego encoder maps the ego state to
\begin{equation}
F_e
=
\mathcal{E}_e(S_{\mathrm{ego}})
\in\mathbb{R}^{T\times1\times D}.
\label{eq:ego_encoding}
\end{equation}
At $t_{\mathrm{cur}}$, a frozen occupancy extractor returns occupied cells
$O^{\mathrm{raw}}=\{(p_i,y_i)\}_{i=1}^{K}$ with ego-centric position $p_i$
and semantic class $y_i$. After range normalization, an MLP projector maps
the cells to occupancy tokens:
\begin{equation}
O_{t_{\mathrm{cur}}}
=
\mathcal{P}_{\mathrm{occ}}(O^{\mathrm{raw}})
\in\mathbb{R}^{K\times D}.
\label{eq:occ_encoding}
\end{equation}
Current visual tokens retrieve relevant occupancy context through
cross-attention,
\begin{equation}
F_{\mathrm{occ}}^{t_{\mathrm{cur}}}
=
\mathcal{A}_{\mathrm{occ}}
\left(
F_v^{t_{\mathrm{cur}}},
O_{t_{\mathrm{cur}}},
O_{t_{\mathrm{cur}}}
\right),
\label{eq:occ_injection}
\end{equation}
where $F_{\mathrm{occ}}^{t_{\mathrm{cur}}}\in\mathbb{R}^{N_v\times D}$.
The planning-time world representation is
\begin{equation}
\mathcal{W}_{t_{\mathrm{cur}}}
=
[F_e^{t_{\mathrm{cur}}};
 F_v^{t_{\mathrm{cur}}};
 F_{\mathrm{occ}}^{t_{\mathrm{cur}}}]
\in\mathbb{R}^{(1+2N_v)\times D}.
\label{eq:world_representation}
\end{equation}
Only current-frame features enter the planner: ego tokens encode navigation
intent and kinematics, visual tokens encode semantics, and occupancy tokens
provide query-selected metric geometry.

\paragraph{Temporal latent prediction.}
During training, the dynamic world model (DWM) predicts future scene-ego latents.
We form $Z_t=[F_e^t;F_v^t]$ and use a Transformer decoder
$\mathcal{D}_{\mathrm{dwm}}$ with 3D rotary positional encoding and a
frame-wise teacher-forcing mask. The positional encoding covers time, camera
view, and query identity, while the mask allows bidirectional within-frame
attention and causal cross-frame attention:
\begin{equation}
\widehat{Z}_{2:T}
=
\mathcal{D}_{\mathrm{dwm}}(Z_{1:T-1}).
\label{eq:dwm_prediction}
\end{equation}
The DWM is optimized with latent prediction and ego-state supervision:
\begin{equation}
\mathcal{L}_{\mathrm{wm}}
=
\MSE(\widehat{Z}_{2:T},Z_{2:T}),
\label{eq:wm_loss}
\end{equation}
\begin{equation}
\mathcal{L}_{\mathrm{ego}}
=
\mathcal{L}_{\mathrm{cmd}}^{\mathrm{ce}}
+\MSE(\widehat{v},v)
+\MSE(\widehat{a},a),
\label{eq:ego_loss}
\end{equation}
where $\mathcal{L}_{\mathrm{cmd}}^{\mathrm{ce}}$ supervises commands and
$(v,a)$ denote ego velocity and acceleration. The total objective is
\begin{equation}
\mathcal{L}_{\mathrm{dwm}}
=
\mathcal{L}_{\mathrm{wm}}
+
\mathcal{L}_{\mathrm{ego}}.
\label{eq:dwm_loss}
\end{equation}

\paragraph{Feature distribution stabilization.}
To prevent temporal prediction from collapsing to low-variance features, we
regularize each visual-token identity across scenes.
Following SIGReg~\citep{balestriero2025lejepa}, let
$f_{b,t,n}=(F_v)_{b,t,n}\in\mathbb{R}^{D}$ be token $n$ of sample $b$ at
frame $t$. For random unit direction $r_j$ and frequency $\omega_\ell$, the
empirical characteristic function is
\begin{equation}
\widehat{\varphi}_{tnj}(\omega_\ell)
=
\frac{1}{B}
\sum_{b=1}^{B}
\exp\!\left(
\mathrm{i}\omega_\ell r_j^\top f_{b,t,n}
\right).
\label{eq:sigreg_characteristic}
\end{equation}
SIGReg matches it to a standard Gaussian:
\begin{equation}
\mathcal{L}_{\mathrm{sigreg}}
=
\frac{B}{T N_v J}
\sum_{t,n,j,\ell}
\widetilde{w}_\ell
\left|
\widehat{\varphi}_{tnj}(\omega_\ell)
-
e^{-\omega_\ell^2/2}
\right|^2,
\label{eq:sigreg}
\end{equation}
where $J$ and $\widetilde{w}_\ell$ denote the number of
random directions and integration weight. This target-free regularizer
maintains a non-degenerate cross-scene distribution of temporally predictive
visual tokens.

\subsection{Trajectory MeanFlow Planner}
\label{sec:trajmf}

The planner first compresses the current world representation into a compact
set of action-conditioned planning tokens. A planning-condition decoder
$\mathcal{D}_{\mathrm{cond}}$ uses $N_a$ learnable action queries
$Q_a\in\mathbb{R}^{N_a\times D}$ to attend to
$\mathcal{W}_{t_{\mathrm{cur}}}$:
\begin{equation}
C_{t_{\mathrm{cur}}}
=
\mathcal{D}_{\mathrm{cond}}
\left(
Q_a,
\mathcal{W}_{t_{\mathrm{cur}}},
\mathcal{W}_{t_{\mathrm{cur}}}
\right)
\in\mathbb{R}^{N_a\times D}.
\label{eq:planning_condition}
\end{equation}
Together, $\mathcal{W}_{t_{\mathrm{cur}}}$ and $C_{t_{\mathrm{cur}}}$ form
the condition $c$ supplied to Trajectory MeanFlow Planner(TrajMF).

Algorithm~\ref{alg:trajmf} summarizes TrajMF training and one-step sampling.
TrajMF formulates planning as conditional flow matching in trajectory
space. We convert the expert trajectory $\tau$ into a normalized relative
trajectory $x=\mathcal{R}(\tau)$ and sample noise $\epsilon$ from a
$K_p$-component Gaussian-mixture prior $p$.
The linear flow path is
\begin{equation}
z_t=(1-t)x+t\epsilon,
\qquad
v_t=\frac{\mathrm{d}z_t}{\mathrm{d}t}=\epsilon-x.
\label{eq:traj_path}
\end{equation}
Specifically, TrajMF directly predicts the denoised trajectory, while applying supervision in the induced velocity space:
\begin{equation}
\widehat{x}_\theta=\mathcal{F}_\theta(z_t,r,t,c).
\label{eq:traj_prediction}
\end{equation}
We convert it to an induced average velocity
\begin{equation}
u_\theta=\frac{z_t-\widehat{x}_\theta}{t}, \qquad t>0,
\label{eq:induced_average_velocity}
\end{equation}
and compute its directional derivative by JVP along the induced trajectory,
with $v_{\mathrm{ins}}=u_\theta(z_t,t,t\mid c)$:
\begin{equation}
(u_\theta,\dot{u}_\theta)
=
\operatorname{JVP}
\left[
u_\theta(\cdot\mid c);(z_t,r,t);(v_{\mathrm{ins}},0,1)
\right].
\label{eq:jvp}
\end{equation}
This avoids explicitly forming the Jacobian. The instantaneous velocity is
\begin{equation}
\widehat{v}_\theta
=
u_\theta+(t-r)\sg(\dot{u}_\theta),
\label{eq:induced_instantaneous_velocity}
\end{equation}
where $\sg(\cdot)$ blocks gradients through the derivative term. The training
loss is
\begin{equation}
\mathcal{L}_{\mathrm{trajmf}}
=
\left\|
\widehat{v}_\theta-v_t
\right\|_1.
\label{eq:trajmf_loss}
\end{equation}

At sampling, no velocity conversion is required. We draw one noise
sample $\epsilon^k$ from each component of $p$ and evaluate
$\mathcal{F}_\theta(\{\epsilon^k\}_{k=1}^{K_p},0,1,c)$ once to generate
$K_p$ multimodal trajectory proposals in a batch. Inside $\mathcal{F}_\theta$,
the conditional Transformer produces a pre-output feature
$h_a^k\in\mathbb{R}^{D}$ for proposal $k$, which the output head
$\mathcal{P}_{\mathrm{out}}$ maps to its trajectory:
\begin{equation}
\widehat{x}_\theta^k
=
\mathcal{P}_{\mathrm{out}}(h_a^k),
H_a=[h_a^1;\ldots;h_a^{K_p}]
\in\mathbb{R}^{K_p\times D}.
\label{eq:action_representation}
\end{equation}
We define $H_a$ as the action representation because it directly supports
trajectory prediction and therefore provides trajectory-related features
for subsequent representation alignment. The proposals
$\{\widehat{x}_\theta^k\}_{k=1}^{K_p}$ are then conditionally aggregated
by $\mathcal{D}_{\mathrm{refine}}$ into $\widehat{x}$. We train $\mathcal{D}_{\mathrm{refine}}$ against the expert
relative trajectory $x$ using
$\mathcal{L}_{\mathrm{traj}}=\|\widehat{x}-x\|_1$.

\input{tables/navsim_v2}
\input{tables/hugsim}

\begin{algorithm}[t]
\caption{TrajMF training and one-step sampling}
\label{alg:trajmf}
\textbf{Input}: trajectory $\tau$, condition $c$, network
$\mathcal{F}_\theta$, refiner $\mathcal{D}_{\mathrm{refine}}$,
trajectory transform $\mathcal{R}$, and multimodal Gaussian prior $p$
with $K_p$ components
\begin{algorithmic}[1]

\STATE \textbf{Training:}
\STATE \hspace{1em} $x\leftarrow\mathcal{R}(\tau)$
\STATE \hspace{1em} Sample $\epsilon\sim p$
\STATE \hspace{1em} Sample $(r,t)$
\STATE \hspace{1em} $z_t\leftarrow(1-t)x+t\epsilon$
\STATE \hspace{1em} $v_t\leftarrow\epsilon-x$

\STATE \hspace{1em} \textbf{function} $u_\theta(z,r,t\mid c)$:
\STATE \hspace{2em}
$\widehat{x}_\theta\leftarrow\mathcal{F}_\theta(z,r,t,c)$
\STATE \hspace{2em}
\textbf{return} $(z-\widehat{x}_\theta)/t$
\STATE \hspace{1em} \textbf{end function}

\STATE \hspace{1em}
$v_{\mathrm{ins}}\leftarrow u_\theta(z_t,t,t\mid c)$
\STATE \hspace{1em}
$(u_\theta,\dot{u}_\theta)
\leftarrow
\operatorname{JVP}
[u_\theta(\cdot\mid c);(z_t,r,t);(v_{\mathrm{ins}},0,1)]$
\STATE \hspace{1em}
$\widehat{v}_\theta
\leftarrow
u_\theta+(t-r)\sg(\dot{u}_\theta)$
\STATE \hspace{1em}
Minimize
$\mathcal{L}_{\mathrm{trajmf}}
\leftarrow
\|\widehat{v}_\theta-v_t\|_1$

\STATE \textbf{Sampling:}
\STATE \hspace{1em}
Sample $\{\epsilon^k\}_{k=1}^{K_p}$ from the $K_p$ components of $p$
\STATE \hspace{1em}
$\{\widehat{x}_\theta^k\}_{k=1}^{K_p}
\leftarrow
\mathcal{F}_\theta
(\{\epsilon^k\}_{k=1}^{K_p},0,1,c)$
\STATE \hspace{1em}
$\widehat{x}
\leftarrow
\mathcal{D}_{\mathrm{refine}}
(\{\widehat{x}_\theta^k\}_{k=1}^{K_p},c)$
\STATE \hspace{1em}
\textbf{return}
$\widehat{\tau}
\leftarrow
\mathcal{R}^{-1}(\widehat{x})$

\end{algorithmic}
\end{algorithm}

\subsection{Visual-Action Representation Alignment}
\label{sec:vara}

World conditioning provides scene context to the action decoder but does
not directly constrain its internal action representation. VARA introduces
a representation-level objective that encourages the action representation
to remain semantically consistent with the current visual representation.
Specifically, we project and average the proposal features:
\begin{equation}
z_a
=
\frac{1}{K_p}
\sum_{k=1}^{K_p}
\mathcal{P}_a(h_a^k),
\label{eq:vara_action}
\end{equation}
and average the current visual tokens with stop-gradient:
\begin{equation}
z_v
=
\sg\!\left(
\frac{1}{N_v}
\sum_{n=1}^{N_v}
F_v^{t_{\mathrm{cur}},n}
\right).
\label{eq:vara_visual}
\end{equation}
Here, $\mathcal{P}_a$ projects the action features into the visual feature
space, and $N_v$ is the number of current visual tokens. The alignment
objective is
\begin{equation}
\mathcal{L}_{\mathrm{align}}
=
1-\operatorname{cos}(z_a,z_v).
\label{eq:vara_loss}
\end{equation}
This cosine objective encourages semantic consistency between the two
pooled representations rather than enforcing element-wise equality. Its
asymmetric gradient design regularizes the action representation while
blocking alignment gradients through the visual target.

\subsection{Training and Inference}
\label{sec:training_inference}

We group the training objectives as
\begin{equation}
\mathcal{L}_{\mathrm{world}}
=
\lambda_1
\left(
\mathcal{L}_{\mathrm{geo}}
+
\mathcal{L}_{\mathrm{dwm}}
+
\mathcal{L}_{\mathrm{sigreg}}
\right),
\label{eq:world_loss}
\end{equation}
\begin{equation}
\mathcal{L}_{\mathrm{action}}
=
\lambda_2\mathcal{L}_{\mathrm{trajmf}}
+
\lambda_3\mathcal{L}_{\mathrm{traj}},
\label{eq:action_loss}
\end{equation}
\begin{equation}
\mathcal{L}
=
\mathcal{L}_{\mathrm{world}}
+
\mathcal{L}_{\mathrm{action}}
+
\lambda_4\mathcal{L}_{\mathrm{align}}.
\label{eq:total_loss}
\end{equation}
We use $\lambda_1=0.2$, $\lambda_2=20.0$, $\lambda_3=7.0$, and
$\lambda_4=2.0$. At inference, only the modality encoders and TrajMF are
used. All inputs are derived from camera images and ego
states, without additional sensor modalities.

%% file: figures/overview_layout.tex
\begin{figure*}[t]
    \centering
    \includegraphics[width=0.94\textwidth]{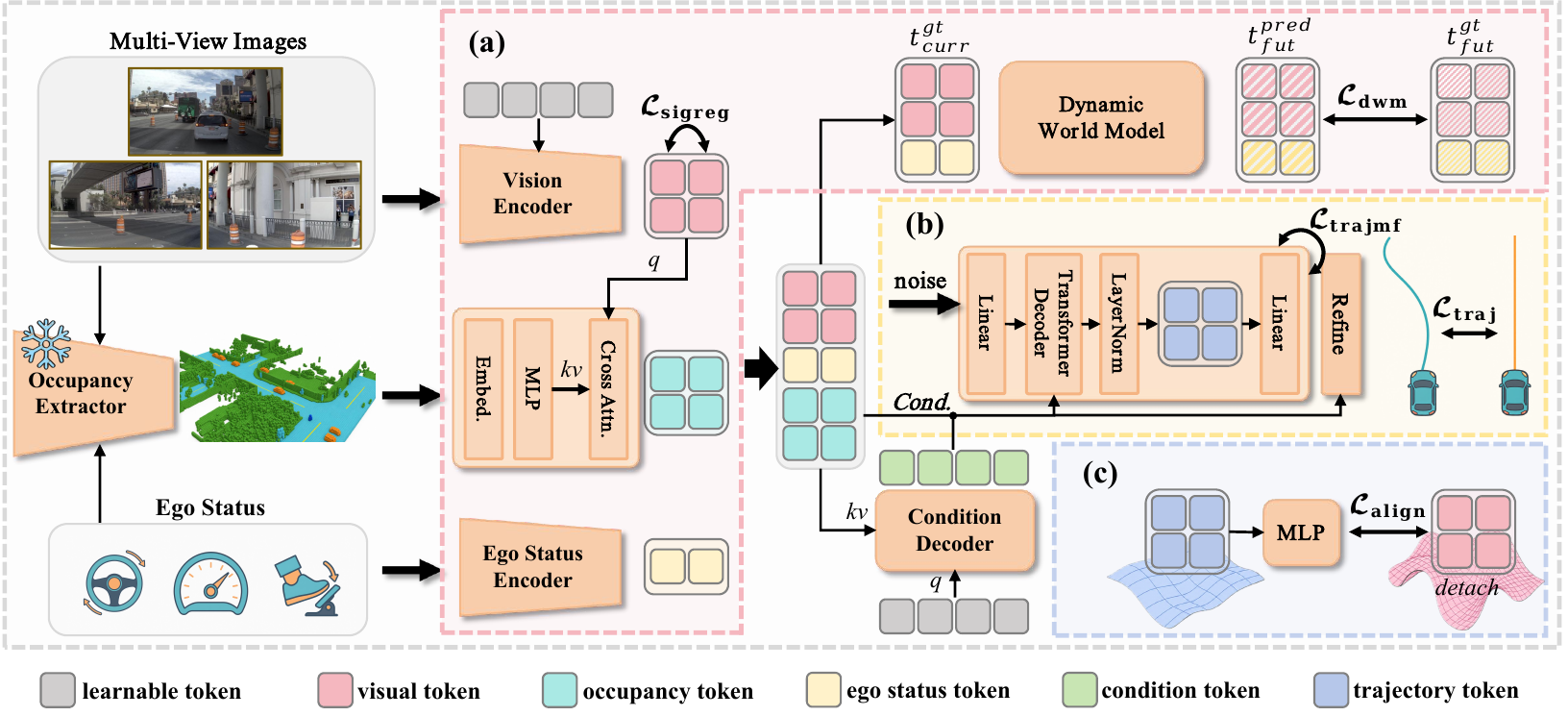}
    \caption{Overview of \method{}.
    (a) WoRC constructs the world representation through scene prediction,
    distribution regularization, and additional geometric context.
    (b) TrajMF generates multimodal trajectory proposals in one step and
    aggregates them into the final trajectory.
    (c) VARA uses stop-gradient visual features to encourage semantic
    consistency between the trajectory-predictive action representation
    and the driving scene.}
    \label{fig:overview}
\end{figure*}

%% file: tables/navsim_v2.tex
\begin{table*}[t]
    \centering
    \setlength{\tabcolsep}{2.2pt}
    \resizebox{0.90\textwidth}{!}{%
    \begin{tabular}{l|ccccccccc|c}
        \toprule
        Method & NC $\uparrow$ & DAC $\uparrow$ & DDC $\uparrow$ & TLC $\uparrow$ & EP $\uparrow$ & TTC $\uparrow$ & LK $\uparrow$ & HC $\uparrow$ & EC $\uparrow$ & \textbf{EPDMS} $\uparrow$ \\
        \midrule
        TransFuser~\citep{transfuser} & 96.9 & 89.9 & 97.8 & \underline{99.7} & 87.1 & 95.4 & 92.7 & \textbf{98.3} & 87.2 & 76.7 \\
        DriveSuprim~\citep{drivesuprim} & 97.5 & 96.5 & 99.4 & 99.6 & \textbf{88.4} & 96.6 & 95.5 & \textbf{98.3} & 77.0 & 83.1 \\
        ReCogDrive~\citep{recogdrive} & 98.3 & 95.2 & 99.5 & \textbf{99.8} & 87.1 & 97.5 & 96.6 & \textbf{98.3} & 86.5 & 83.6 \\
        DiffusionDrive~\citep{diffusiondrive} & 98.2 & 95.9 & 99.4 & \textbf{99.8} & 87.5 & 97.3 & 96.8 & \textbf{98.3} & 87.7 & 84.5 \\
        World4Drive~\citep{world4drive} & 97.8 & 96.3 & 99.4 & \textbf{99.8} & \underline{88.3} & 97.1 & \textbf{97.7} & 98.0 & 53.9 & 84.8 \\
        DriveVLA-W0~\citep{drivevlaw0} & \textbf{98.5} & \textbf{99.1} & 98.0 & \underline{99.7} & 86.4 & \underline{98.1} & 93.2 & 97.9 & 58.9 & 86.1 \\
        Drive-JEPA~\citep{drivejepa} & \underline{98.4} & \underline{98.6} & 99.1 & \textbf{99.8} & \textbf{88.4} & 97.8 & \underline{97.6} & 97.9 & 84.8 & 87.8 \\
        MeanFuser~\citep{meanfuser} & 98.3 & 97.2 & \underline{99.6} & \textbf{99.8} & 87.6 & 97.4 & 97.3 & \textbf{98.3} & \underline{88.2} & \underline{89.5} \\
        \midrule
        Latent-WAM~\citep{latentwam} & 98.1 & 97.3 & \underline{99.6} & \textbf{99.8} & 87.7 & 97.3 & \underline{97.6} & \underline{98.1} & 87.3 & 89.3 \\
        \textbf{\method} & \textbf{98.5} & 97.5 & \textbf{99.7} & \textbf{99.8} & 87.8 & \textbf{98.3} & \textbf{97.7} & \textbf{98.3} & \textbf{89.0} & \textbf{90.1} \\
        \bottomrule
    \end{tabular}%
    }
    \caption{Comparison on NAVSIM v2~\citep{navsim,navsimv2metrics}. Best and second-best results are bold and underlined.}
    \label{tab:navsim_v2}
\end{table*}

%% file: tables/hugsim.tex
\begin{table*}[t]
    \centering
    \setlength{\tabcolsep}{2.2pt}
    \resizebox{0.90\textwidth}{!}{%
    \begin{tabular*}{\textwidth}{@{\extracolsep{\fill}}lcccccccccc@{}}
        \toprule
        & \multicolumn{5}{c}{RC $\uparrow$} & \multicolumn{5}{c}{HD-Score $\uparrow$} \\
        \cmidrule(lr){2-6}\cmidrule(l){7-11}
        Method & E & M & H & X & \textbf{Avg.} & E & M & H & X & \textbf{Avg.} \\
        \midrule
        UniAD~\citep{uniad} & 58.6 & 41.2 & \textbf{40.4} & 26.0 & 40.6 & 48.7 & \underline{29.5} & \textbf{27.3} & \underline{14.3} & \underline{28.9} \\
        VAD~\citep{vad} & 38.7 & 27.0 & 25.5 & 23.0 & 27.9 & 24.3 & 9.9 & 10.4 & 8.2 & 12.3 \\
        LTF~\citep{transfuser} & 68.4 & 40.7 & 36.9 & 25.5 & 41.4 & 52.8 & 24.6 & 19.8 & 8.1 & 24.8 \\
        GTRS-Dense\textsuperscript{$\dagger$}~\citep{gtrs} & 64.2 & \textbf{50.0} & 20.7 & 22.3 & 38.0 & 55.5 & \textbf{39.0} & 11.7 & \underline{14.3} & 28.6 \\
        MeanFuser~\citep{meanfuser} & 76.7 & 41.1 & 37.9 & 28.4 & 43.9 & 67.1 & 24.3 & \underline{21.5} & 11.0 & 28.4 \\
        \midrule
        Latent-WAM~\citep{latentwam} & \textbf{84.2} & 42.5 & 30.6 & \textbf{35.5} & \underline{45.9} & \textbf{72.5} & 24.0 & 12.2 & \textbf{18.1} & \underline{28.9} \\
        \textbf{\method} & \underline{80.1} & \underline{45.1} & \underline{38.7} & \underline{32.0} & \textbf{47.0} & \underline{68.9} & 26.8 & 20.5 & 13.9 & \textbf{30.1} \\
        \bottomrule
    \end{tabular*}
    }
    \caption{Zero-shot closed-loop evaluation on HUGSIM~\citep{hugsim} across easy (E), medium (M), hard (H), and extreme (X) settings. $\dagger$: GTRS-Dense is evaluated on 345 scenarios.}
    \label{tab:hugsim}
\end{table*}

%% file: src/04_experiments.tex
\section{Experiments}\label{sec:experiments}
\input{figures/stabilizer_layout}
\input{figures/traj_compare_layout}
\input{figures/align_feat_layout}
\input{tables/ablation}
\subsection{Experimental Setup}\label{sec:experimental_setup}
\paragraph{Benchmarks.}
NAVSIM~\citep{navsim} is a large-scale real-world driving benchmark with
pseudo-closed-loop evaluation, comprising approximately 103k training and
12k evaluation scenarios. We train on \texttt{navtrain} and evaluate on
\texttt{navtest} using Extended Predictive Driver Model Score (EPDMS).
The original PDMS comprises No At-Fault Collisions (NC), Drivable Area
Compliance (DAC), Time to Collision (TTC), Ego Progress (EP), and Comfort
(C). NAVSIM v2~\citep{navsimv2metrics} extends it with Driving Direction
Compliance (DDC), Traffic Light Compliance (TLC), and Lane Keeping (LK),
and refines comfort into History Comfort (HC) and Extended Comfort (EC).

We further evaluate the same NAVSIM-trained model without fine-tuning in
HUGSIM~\citep{hugsim}, a photorealistic closed-loop simulator constructed
from diverse real-world driving scenarios. We report Route Completion (RC)
and HD-Score across four difficulty levels. HD-Score combines route
completion with per-frame safety, drivable-area compliance, collision
risk, and comfort.

\paragraph{Implementation.}
We use DINOv2-Base~\citep{dinov2} as the visual encoder with three
$224\times448$ camera views and 16 scene queries per view. All latent
features are projected to a shared dimension of 256. A frozen occupancy
extractor processes the current images and ego state, while a frozen
image-based geometry teacher provides geometric distillation during
training. We train the model with Adam for 25 epochs using a total batch size of 128. The learning rate is linearly warmed to
$10^{-4}$ over the first 10\% of steps and then cosine-decayed to
$10^{-6}$. End-to-end training of the planner takes approximately 12 hours on 16 NVIDIA H20 GPUs.

\subsection{Main Results}\label{sec:main_results}
\paragraph{NAVSIM v2.}
As shown in Table~\ref{tab:navsim_v2}, \method{}
achieves the highest EPDMS of 90.1 among all compared methods, including
those using additional sensor modalities, and outperforms Latent-WAM by 0.8. Except for DAC and EP, our model obtains the
best or tied-best results on all component metrics, demonstrating balanced
performance in safety, rule compliance, lane keeping, and comfort.
Notably, it achieves 89.0 EC, substantially surpassing prior methods and
highlighting its strong capability for comfortable driving.

\paragraph{Zero-shot HUGSIM.}
Table~\ref{tab:hugsim} reports zero-shot closed-loop evaluation without
HUGSIM fine-tuning. \method{} achieves weighted averages of 47.0 RC and
30.1 HD-Score, the highest reported values among existing methods,
outperforming Latent-WAM by 1.1 and 1.2
points, respectively. It also ranks among the top two methods in RC across
all four difficulty levels, demonstrating robust route completion from easy
to extreme scenarios. The improved HD-Score further indicates stronger
overall driving quality in safety, compliance, and comfort.

\subsection{Ablation Studies}\label{sec:ablations}
\input{tables/action_decoder}

\paragraph{World representation.}
Configurations 1--4 in Table~\ref{tab:ablation} compare stabilization under the same MLP planner. Dynamic world model(DWM) alone obtains 88.0 versus the 88.2 baseline. EMA mitigates collapse and reaches 88.6 but adds an 86.6M-parameter target encoder; SIGReg reaches 88.8 without that branch. Occupancy further raises performance to 89.1. Thus temporal prediction benefits from stabilization, with SIGReg providing the better distribution statistics and lower training parameter count in our setting.

\paragraph{Action representation.}
With the world representation fixed, Table~\ref{tab:action_decoder}
compares action decoders. MLP reaches 89.1 EPDMS, while Improved MeanFlow
(iMF) and TrajMF achieve comparable scores of 89.4 and 89.5. This result
is consistent with Pixel MeanFlow~\citep{pixelmeanflow}: in a
low-dimensional trajectory space, both average-velocity and trajectory
prediction can be effective. However, TrajMF reaches 89.5 at 8,080
iterations, whereas iMF requires 16,160 iterations to reach 89.4,
demonstrating comparable performance with fewer optimization steps.

Adding VARA to TrajMF raises EPDMS from 89.5 to 89.8; geometry
distillation further improves it to 90.1, indicating that geometric
features benefit trajectory planning. Overall, the ablation
results demonstrate the effectiveness and complementary gains of all components.

\subsection{More Analysis}
\label{sec:representation_analysis}

\paragraph{Stabilizing the visual-token distribution.}
DWM predicts future latent features produced by the trainable
visual encoder. Without an explicit constraint on the feature
distribution, this self-supervised objective admits a low-variance
solution: mapping different scenes to similar tokens reduces the
prediction loss without preserving scene-specific information.
Figure~\ref{fig:stabilizer} quantifies this effect: for each token
identity, (a) reports the feature standard deviation across
scenes, averaged over channels, while (b) shows the empirical CDF
of same-token cosine similarities across scenes. Curves rising near one
indicate nearly identical features, whereas curves rising near zero
indicate lower similarity. Without stabilization, the mean feature
standard deviation is 0.088 and the mean cosine similarity is 0.973.
EMA partially alleviates this behavior, whereas SIGReg changes these
statistics to 0.900 and 0.017, respectively. Thus, SIGReg avoids the
low-variance solution without an additional target encoder.

\paragraph{Aligning visual and action representations.}
Figure~\ref{fig:align_feat} visualizes action representations with and
without VARA. Colors denote $K=8$ clusters independently derived from
view-specific CLS tokens of a frozen DINOv2~\citep{dinov2}, which is
not involved in planner training or VARA supervision. VARA yields more
coherent neighborhoods under this external visual reference. Together
with the EPDMS improvement from 89.5 to 89.8, these results demonstrate
the effectiveness of VARA.

\paragraph{Qualitative analysis of trajectories.}
Figure~\ref{fig:traj_compare} compares ours with
World4Drive and Latent-WAM.
Across diverse road conditions, our trajectories (yellow) more closely
follow human trajectories (green), demonstrating better planning.

%% file: figures/stabilizer_layout.tex
\begin{figure}[!htb]
    \centering
    \includegraphics[width=0.80\columnwidth]{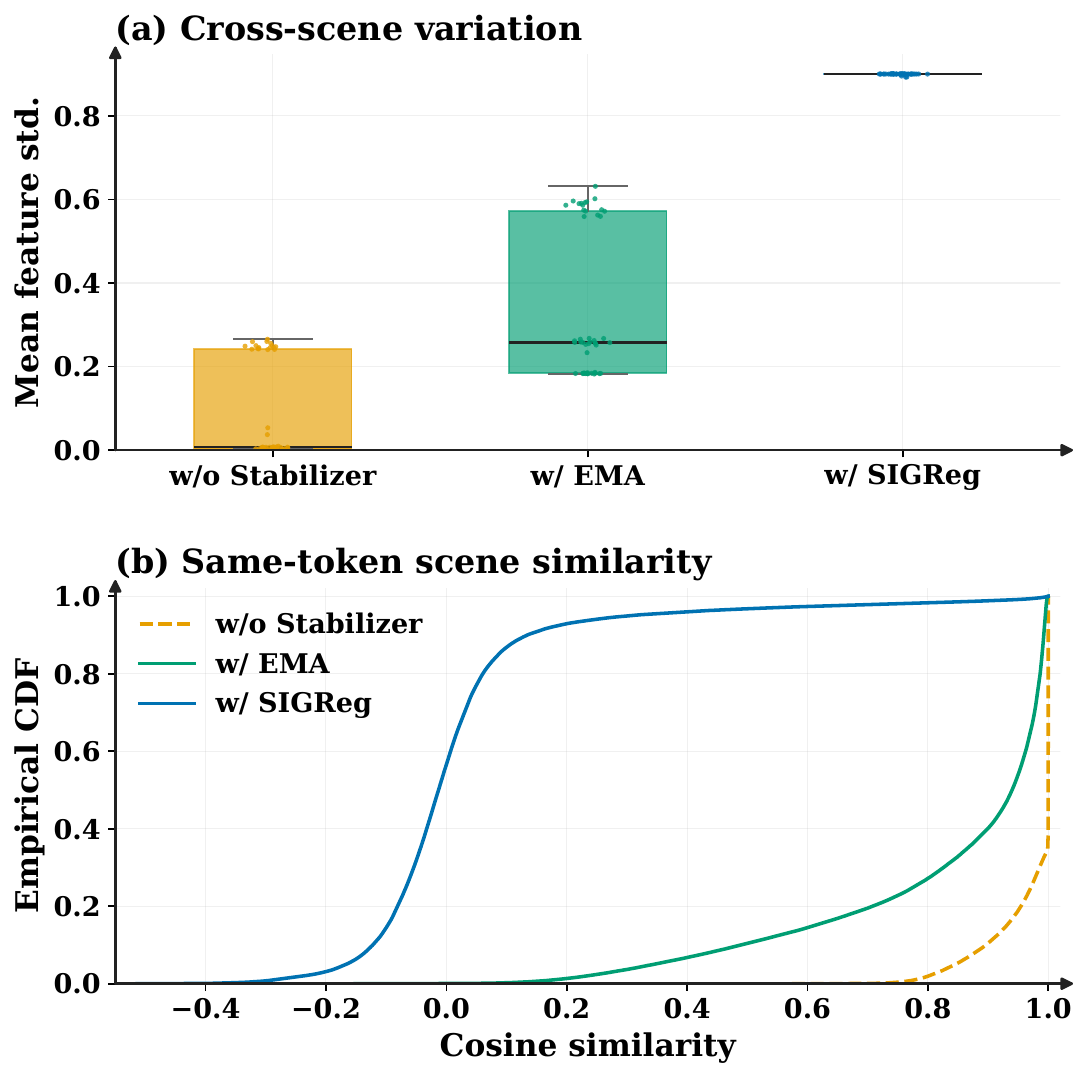}
    \vspace{-1mm}
    \caption{
        Cross-scene visual-token statistics without stabilization, with EMA,
        and with SIGReg. (a) Mean feature standard deviation per token identity;
        (b) Empirical CDF of same-token cosine similarities across scenes. SIGReg produces
        the largest standard deviation and lowest similarity.
    }
    \label{fig:stabilizer}
\end{figure}

%% file: figures/traj_compare_layout.tex
\begin{figure*}[!t]
    \centering
    \includegraphics[width=0.91\textwidth]{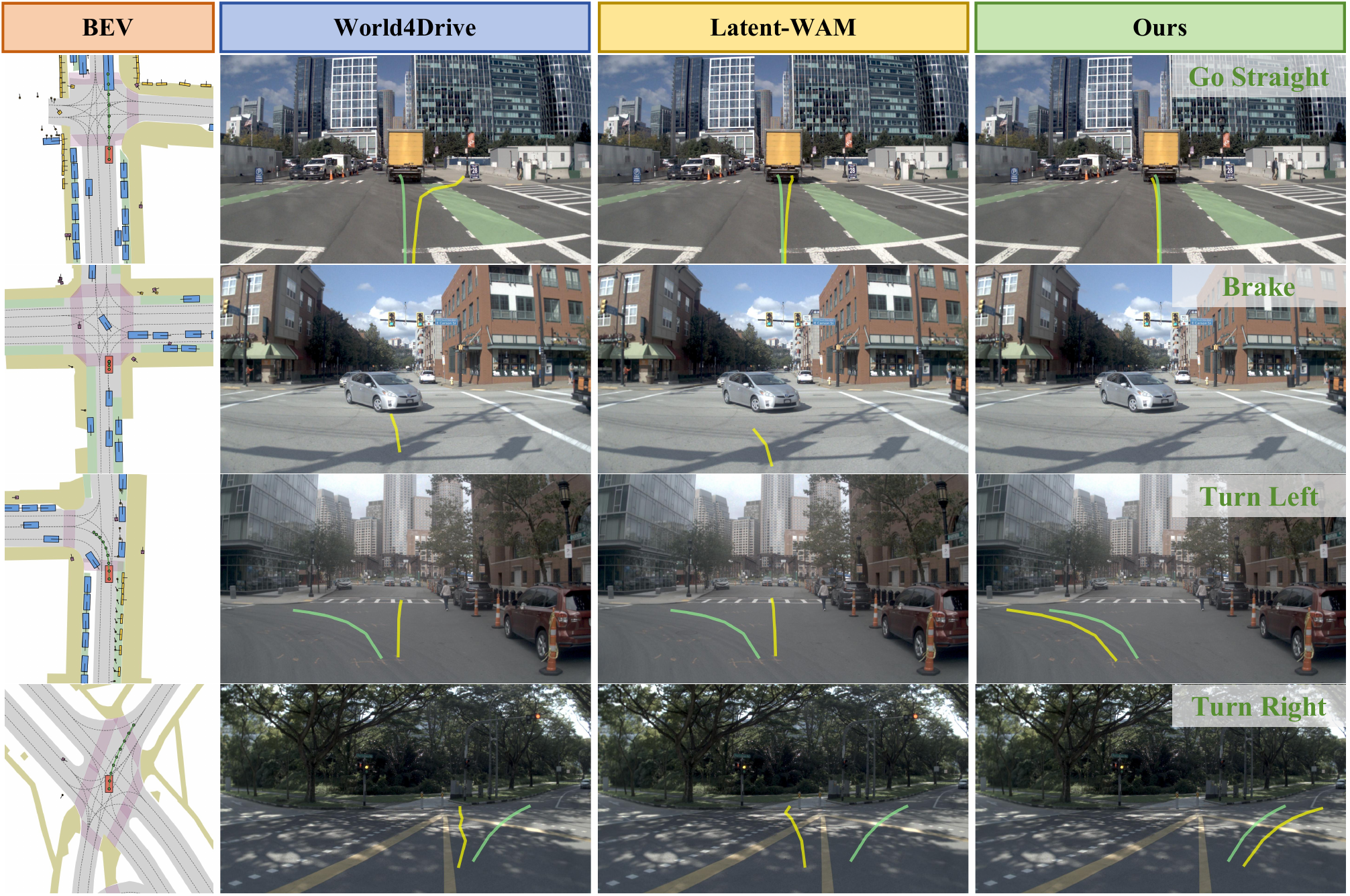}
    \caption{NAVSIM v2 trajectory comparison. Yellow and green denote model and human-driver trajectories. \method{} follows the reference more consistently than World4Drive and Latent-WAM in the displayed scenes.}
    \label{fig:traj_compare}
\end{figure*}

%% file: figures/align_feat_layout.tex
\begin{figure}[!tb]
    \centering
    \includegraphics[width=0.90\columnwidth]{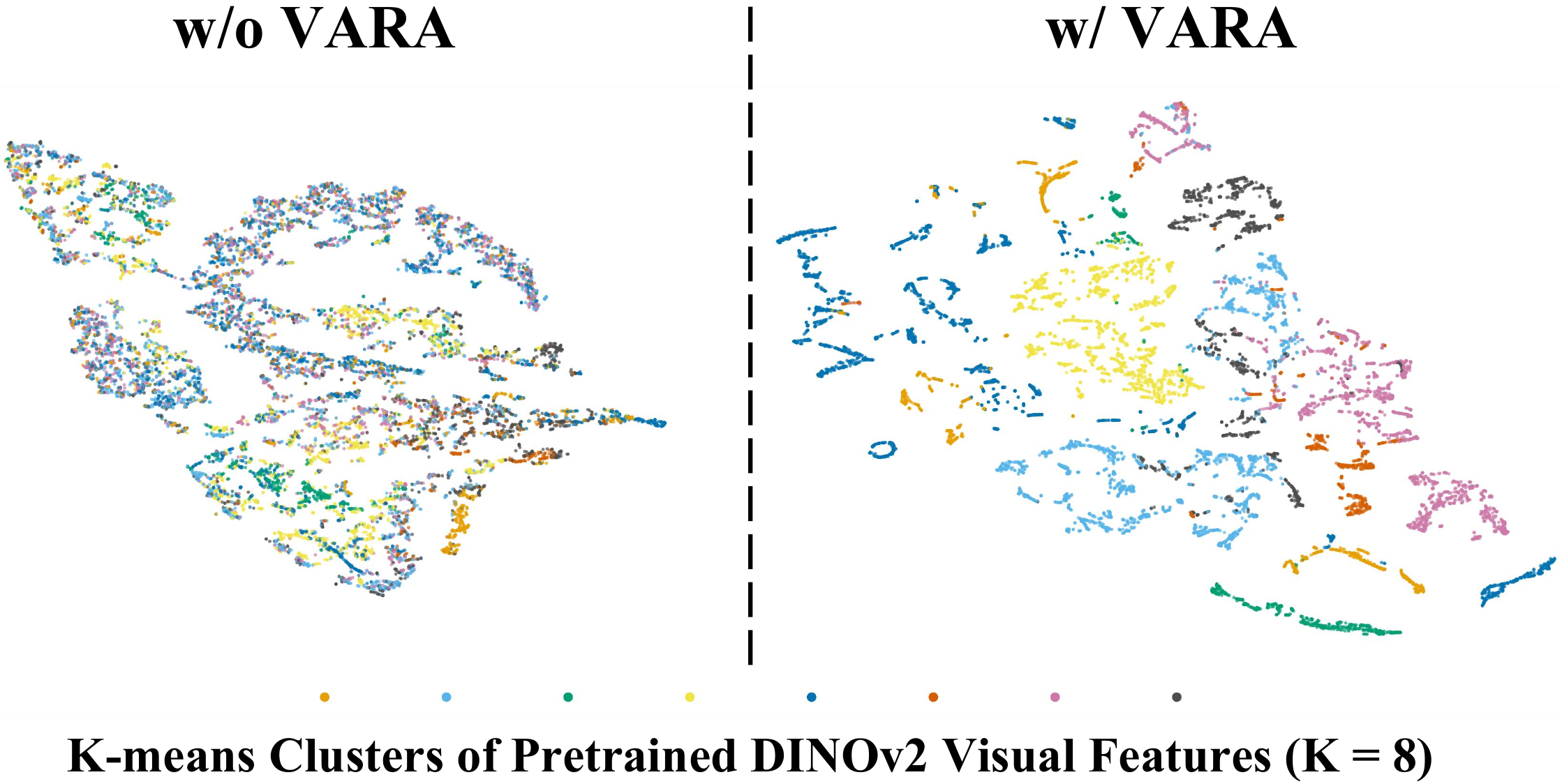}
    \caption{Action representations w/o and w/ VARA. Colors are $K=8$ clusters from frozen DINOv2 CLS features.}
    \label{fig:align_feat}
\end{figure}

%% file: tables/ablation.tex
\begin{table*}[!t]
    \centering
    \begin{tabular*}{\textwidth}{@{\extracolsep{\fill}}c|cccccc|cc@{}}
        \toprule
        ID & DWM & Stabilizer & Occ & Geometry Distill. & Decoder & VARA & EPDMS $\uparrow$ & $\Delta$ EPDMS \\
        \midrule
        1 & \nomark & \nomark & \nomark & \nomark & MLP & \nomark & 88.2 & \tbd \\
        2 & \yesmark & \nomark & \nomark & \nomark & MLP & \nomark & 88.0 & $-0.2$ \\
        3 & \yesmark & EMA$_{\mbox{\fontsize{8}{9}\selectfont +86.6M}}$ & \nomark & \nomark & MLP & \nomark & 88.6 & $+0.4$ \\
        4 & \yesmark & SIGReg & \nomark & \nomark & MLP & \nomark & 88.8 & $+0.6$ \\
        5 & \yesmark & SIGReg & \yesmark & \nomark & MLP & \nomark & 89.1 & $+0.9$ \\
        \midrule
        6 & \yesmark & SIGReg & \yesmark & \nomark & TrajMF & \nomark & 89.5 & $+1.3$ \\
        7 & \yesmark & SIGReg & \yesmark & \nomark & TrajMF & \yesmark & 89.8 & $+1.6$ \\
        8 & \yesmark & SIGReg & \yesmark & \yesmark & TrajMF & \yesmark & \textbf{90.1} & $\mathbf{+1.9}$ \\
        \bottomrule
    \end{tabular*}
    \caption{Unified ablation of world-representation learning, action decoding, and geometric supervision. IDs index configurations; the EMA subscript reports additional parameters, and Geometry Distill. denotes feature alignment to the frozen geometry teacher. $\Delta$ EPDMS is measured relative to configuration 1 (88.2).}
    \label{tab:ablation}
\end{table*}

%% file: tables/action_decoder.tex
\begin{table}[!htb]
    \centering
    \setlength{\tabcolsep}{3pt}
    \begin{tabular}{@{}l|cc|cc@{}}
        \toprule
        Decoder & Pred. & Loss
        & EPDMS $\uparrow$
        & Iter. $\downarrow$ \\
        \midrule
        MLP
        & $x$ & $x$
        & 89.1 & 16,160 \\
        iMF
        & $u$ & $v$
        & 89.4 & 16,160 \\
        TrajMF
        & $x$ & $v$
        & 89.5 & \textbf{8,080} \\
        \bottomrule
    \end{tabular}

    \begin{minipage}{\columnwidth}
        $x$:trajectory, $u$:average
        velocity, $v$:instantaneous velocity.
    \end{minipage}

    \caption{Action-decoder ablation under the same training recipe with
    all other modules fixed. iMF and TrajMF differ only in their network
    prediction targets: iMF predicts $u$, whereas TrajMF predicts $x$.
    Iter. denotes the training iteration attaining the reported EPDMS.}
    \label{tab:action_decoder}
\end{table}

%% file: src/05_conclusion.tex
\section{Conclusion}
\label{sec:conclusion}

Latent-WAM++ improves the representations that link world modeling and
action generation. WoRC learns predictive, stable, and geometry-aware
world features; TrajMF predicts multiple possible trajectories with a
velocity-based objective; and VARA aligns action features with visual
semantics. Latent-WAM++ achieves state-of-the-art results on NAVSIM v2
and zero-shot HUGSIM at a lower end-to-end training cost than Latent-WAM.
These results show the value of designing each representation for its
role in planning and linking them through conditional generation and
feature alignment.

%% file: appendix.tex
\begingroup
\renewcommand{\yesmark}{\ding{51}}
\renewcommand{\nomark}{\ding{55}}
\setlength{\emergencystretch}{2em}
\widowpenalty=10000
\clubpenalty=10000
\displaywidowpenalty=10000

\clearpage

{
\begin{center}
    \textbf{\LARGE Supplementary Material}
\end{center}
}

\suppressfloats[t]
This supplementary material presents the datasets and evaluation
protocols, followed by additional qualitative trajectory results on
NAVSIM and HUGSIM.

\section{Datasets and Evaluation Protocols}
\label{app:sec:benchmark_details}

\subsection{NAVSIM v2}
\label{app:sec:navsim_v2_supp}

\paragraph{Dataset and evaluation protocol.}
NAVSIM~\citep{navsim} is a large-scale autonomous-driving benchmark
built on OpenScene and nuPlan data.
It contains approximately 103\,K training scenarios and 12\,K
evaluation scenarios collected from diverse real-world driving
conditions.
We train our model on the \texttt{navtrain} split and evaluate on
\texttt{navtest} using the Extended Predictive Driver Model Score
(EPDMS) introduced in NAVSIM v2~\citep{navsimv2}.
Our evaluation uses a reactive, pseudo-closed-loop simulator to
estimate how a predicted ego trajectory would interact with the
recorded scene and traffic participants.
For \name{}, the planner receives three RGB camera views and ego
status as its online inputs.

\paragraph{Extended Predictive Driver Model Score.}
NAVSIM v1 evaluates trajectory planning using the Predictive Driver
Model Score (PDMS), which combines safety, drivable-area compliance,
progress, and comfort.
NAVSIM v2 extends this metric to the Extended Predictive Driver Model
Score (EPDMS). Under the official definition, every component is
filtered against its corresponding human-driver score:
\begin{equation}
\widetilde{m}
=
\operatorname{filter}_{m}(\mathrm{agent},\mathrm{human})
=
\begin{cases}
1, & m(\mathrm{human})=0,\\
m(\mathrm{agent}), & \text{otherwise}.
\end{cases}
\end{equation}
When all components are available, the score is
\begin{equation}
\label{app:eq:epdms}
\mathrm{EPDMS}
=
\widetilde{\mathrm{NC}}\cdot
\widetilde{\mathrm{DAC}}\cdot
\widetilde{\mathrm{DDC}}\cdot
\widetilde{\mathrm{TLC}}\cdot
\frac{
5(\widetilde{\mathrm{EP}}+\widetilde{\mathrm{TTC}})
+2(\widetilde{\mathrm{LK}}+\widetilde{\mathrm{HC}}+
\widetilde{\mathrm{EC}})
}{16}.
\end{equation}
Here, NC, DAC, DDC, and TLC denote No At-Fault Collision,
Drivable Area Compliance, Driving Direction Compliance, and Traffic
Light Compliance, respectively.
EP and TTC denote Ego Progress and Time to Collision, while LK, HC,
and EC denote Lane Keeping, History Comfort, and Extended Comfort.
Compared with PDMS, NAVSIM v2 introduces DDC, TLC, and LK to measure
rule and lane compliance, and replaces the original comfort term with
the more detailed HC and EC metrics.
The human-based filter suppresses false-positive penalties when the
recorded human trajectory incurs the same violation. The multiplicative
terms in Eq.~\eqref{app:eq:epdms} then act as safety and rule-compliance
gates, whereas the remaining terms measure progress, collision margin,
lane keeping, and comfort.
When extended comfort cannot be computed for a scene, its term and
weight are omitted and the remaining weights are renormalized.
We report all nine component metrics together with the aggregate
EPDMS in the main paper to provide a complete view of planning
performance across all evaluated criteria.

\subsection{HUGSIM}
\label{app:sec:hugsim_supp}

\paragraph{Dataset and zero-shot protocol.}
HUGSIM~\citep{hugsim} is a photorealistic closed-loop simulator
constructed from real-world scenes in KITTI-360, nuScenes, PandaSet,
and Waymo.
The captured scenes are reconstructed as three-dimensional
environments in which the planner controls the ego vehicle through
RGB observations rendered from dynamically updated viewpoints.
Consequently, the predicted trajectory affects subsequent
observations, enabling reactive closed-loop evaluation.

We use the pre-challenge HUGSIM test set, comprising 80 easy (E),
157 medium (M), 96 hard (H), and 103 extreme (X) scenarios, for a
total of 436.
We directly evaluate the checkpoint trained on NAVSIM v2 on HUGSIM
in a zero-shot manner, without any training or fine-tuning on HUGSIM.
As reported in the main paper, this protocol yields weighted averages
of 47.0 RC and 30.1 HD-Score.
We report Route Completion (RC) and the HUGSIM Driving Score
(HD-Score) for each difficulty level.
For an evaluation metric $m$, the overall average is weighted by the
number of scenarios in each difficulty subset:
\begin{equation}
m_{\mathrm{avg}}
=
\frac{\sum_{d\in\{\mathrm{E,M,H,X}\}}N_d\,m_d}
{\sum_{d\in\{\mathrm{E,M,H,X}\}}N_d}.
\end{equation}
Here, $N_d$ denotes the number of scenarios in difficulty subset $d$,
while $m_d$ is the corresponding subset-level metric value used in the
weighted average.

\paragraph{HUGSIM Driving Score.}
At evaluation timestamp $t$, HUGSIM first computes
\begin{equation}
\mathrm{HD\mbox{-}Score}_{t}
=
\mathrm{NC}_{t}\cdot
\mathrm{DAC}_{t}\cdot
\frac{
5\,\mathrm{TTC}_{t}
+2\,\mathrm{COM}_{t}
}{7},
\end{equation}
where NC, DAC, TTC, and COM denote No Collision,
Drivable Area Compliance, Time to Collision, and Ego Comfort,
respectively.
The final route-level score averages the per-timestamp driving scores
and scales them by the route completion ratio $R_c\in[0,1]$:
\begin{equation}
\mathrm{HD\mbox{-}Score}
=
R_c\cdot
\frac{1}{T}
\sum_{t=1}^{T}
\mathrm{HD\mbox{-}Score}_{t}.
\end{equation}
Thus, a method must both complete the route and maintain safe,
compliant, and comfortable behavior to obtain a high HD-Score.
Following the official implementation, NC and TTC also account for
collisions with static background entities, such as buildings,
fences, and vegetation, using semantic segmentation.

\subsection{NAVSIM-Occ}
\label{app:sec:navsim_occ}

NAVSIM provides RGB images, LiDAR sweeps, three-dimensional object
boxes, and vectorized BEV maps, but it does not provide semantic
occupancy labels for every frame used by the planning benchmark.
Because NAVSIM is derived from the OpenScene redistribution of
nuPlan, we build \textbf{NAVSIM-Occ} upon the NuPlan-Occ annotations
introduced by UniSceneV2~\citep{uniscenev2}.
Following the NuPlan-Occ construction and cleaning protocol, we clean
the available occupancy annotations, align them with the corresponding
NAVSIM frame tokens and sensor calibration, and supplement the missing
frames.
This procedure produces a complete semantic-occupancy extension that
is aligned one-to-one with the NAVSIM scenes, rather than an independent
driving dataset.
NAVSIM-Occ contains 103,288 training scenes and 12,146 test scenes.
Only the training split is used to optimize the occupancy extractor;
the test annotations are reserved for occupancy evaluation.

The resulting occupancy volume covers
$x,y\in[-50,50]$ m and $z\in[-3,5]$ m at a native resolution of
$0.25$ m, resulting in a $400\times400\times32$ voxel grid.
Each occupied voxel is associated with one of eight semantic classes:
vehicle, bicycle, pedestrian, traffic cone, barrier, construction-zone
sign, generic object, and background; unoccupied voxels are assigned
the additional \emph{free} class.
Tab.~\ref{app:tab:navsim_occ_statistics} summarizes the semantic-class
distribution of the occupied voxels. The training and test splits contain
44,824,161,036 and 5,446,767,582 occupied voxels, respectively.
The distribution is strongly imbalanced: background accounts for
92.6545\% of the training split and 94.8239\% of the test split. Here,
\texttt{background} denotes occupied voxels assigned to the non-object
background category and is therefore distinct from the \emph{free}
class used for unoccupied voxels.

\begin{table}[H]
    \centering
    \small
    \setlength{\tabcolsep}{5pt}
    \begin{tabular}{@{}lrrrr@{}}
        \toprule
        & \multicolumn{2}{c}{Training split}
        & \multicolumn{2}{c}{Test split} \\
        \cmidrule(lr){2-3}\cmidrule(l){4-5}
        Semantic class
        & Voxel count & Share (\%)
        & Voxel count & Share (\%) \\
        \midrule
        \texttt{background}
        & 41,531,592,449 & 92.6545
        & 5,164,837,360 & 94.8239 \\
        \texttt{vehicle}
        & 2,899,332,761 & 6.4682
        & 255,186,420 & 4.6851 \\
        \texttt{pedestrian}
        & 208,170,168 & 0.4644
        & 12,563,558 & 0.2307 \\
        \texttt{generic\_object}
        & 143,724,597 & 0.3206
        & 8,735,890 & 0.1604 \\
        \texttt{traffic\_cone}
        & 18,276,533 & 0.0408
        & 1,810,800 & 0.0332 \\
        \texttt{barrier}
        & 16,868,356 & 0.0376
        & 2,959,172 & 0.0543 \\
        \texttt{bicycle}
        & 4,991,342 & 0.0111
        & 542,573 & 0.0100 \\
        \texttt{czone\_sign}
        & 1,204,830 & 0.0027
        & 131,809 & 0.0024 \\
        \midrule
        \textbf{Total}
        & \textbf{44,824,161,036} & \textbf{100.0000}
        & \textbf{5,446,767,582} & \textbf{100.0000} \\
        \bottomrule
    \end{tabular}
    \caption{Semantic-class distribution of occupied NAVSIM-Occ voxels.
    Percentages are computed within each split; \emph{free} denotes
    unoccupied space.}
    \label{app:tab:navsim_occ_statistics}
\end{table}

For occupancy-extractor training, we apply majority pooling within
each $2\times2\times2$ voxel block, yielding a
$200\times200\times16$ grid whose cells have a physical size of
$0.5\times0.5\times0.5$ m for occupancy-extractor supervision.

Figure~\ref{app:fig:navsim_occ_visualization} visualizes the semantic
occupancy annotations projected onto the left, front, and right camera views.

\begin{figure}[H]
    \centering
    \includegraphics[width=0.86\textwidth]{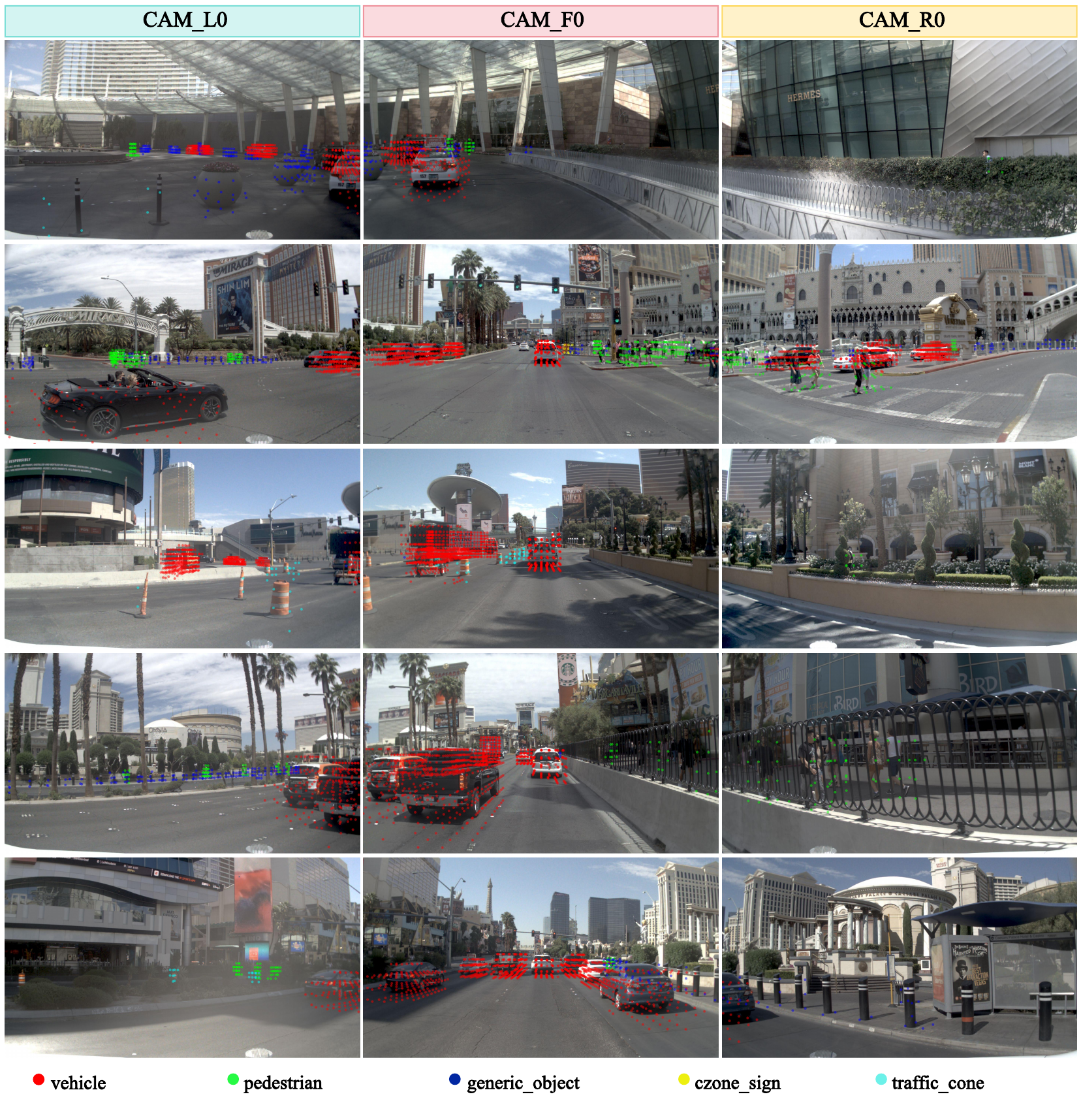}
    \caption{Projected NAVSIM-Occ semantic occupancy annotations for the
    left, front, and right camera views; colors identify classes.}
    \label{app:fig:navsim_occ_visualization}
\end{figure}

\clearpage
\section{Additional Qualitative Results}
\label{app:sec:additional_qualitative_results}

\subsection{Additional NAVSIM Qualitative Results}
\label{app:sec:navsim_qualitative_results}

\begin{figure}[H]
    \centering
    \includegraphics[width=0.92\textwidth,height=0.84\textheight,keepaspectratio]{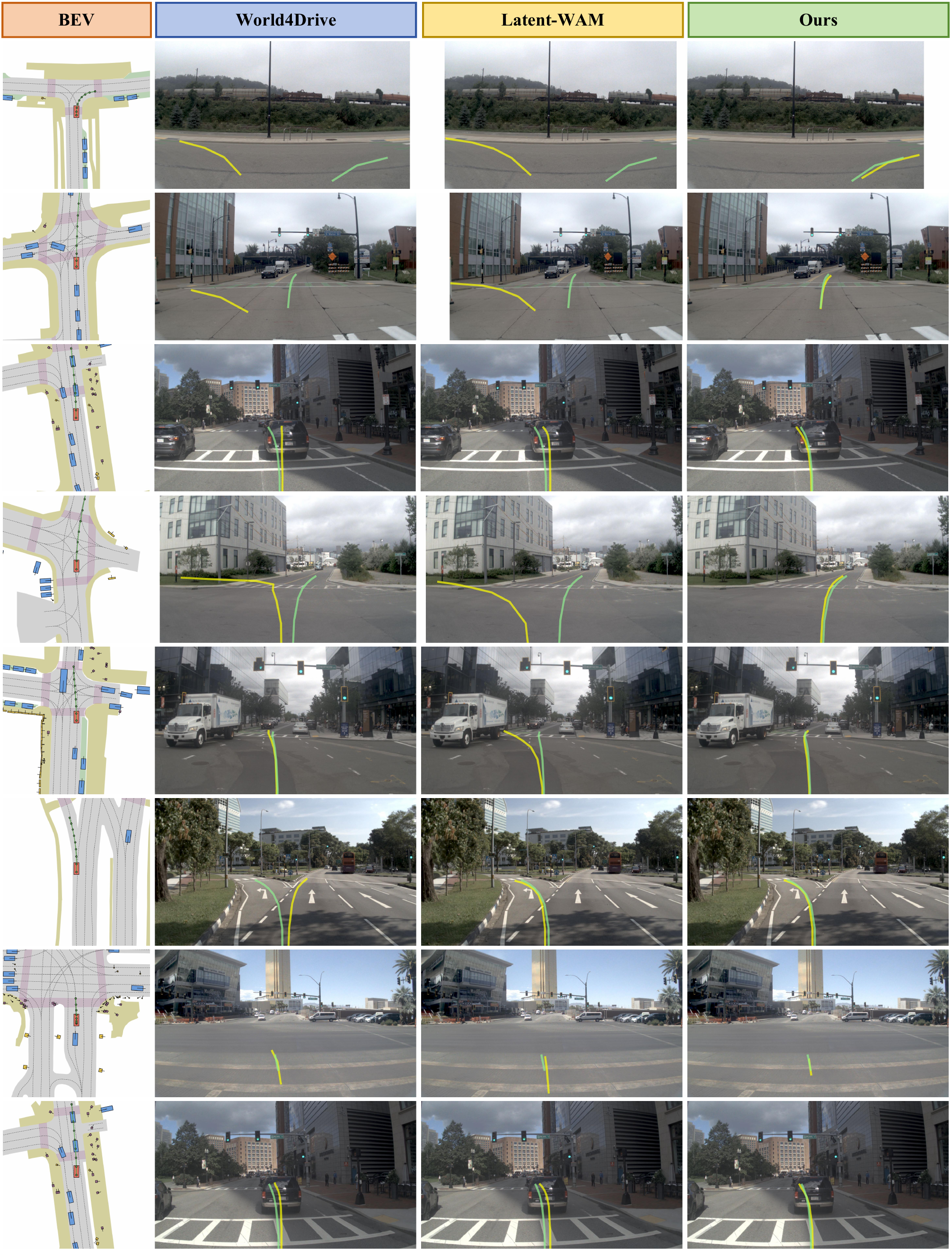}
    \caption{Qualitative trajectory comparison among World4Drive~\citep{world4drive},
    Latent-WAM, and our method. Yellow curves denote model-planned
    trajectories, whereas green curves denote human-driver trajectories.}
    \label{app:fig:trajectory_comparison}
\end{figure}

\clearpage
\subsection{Additional HUGSIM Qualitative Results}
\label{app:sec:hugsim_qualitative_results}

We provide additional trajectory-planning visualizations on HUGSIM.
The four examples use reconstructed scenes and are named by their
source datasets.

\begin{figure}[H]
    \centering
    \begin{subfigure}[t]{0.44\textwidth}
        \centering
        \includegraphics[width=\linewidth,trim=180 190 135 165,clip]
        {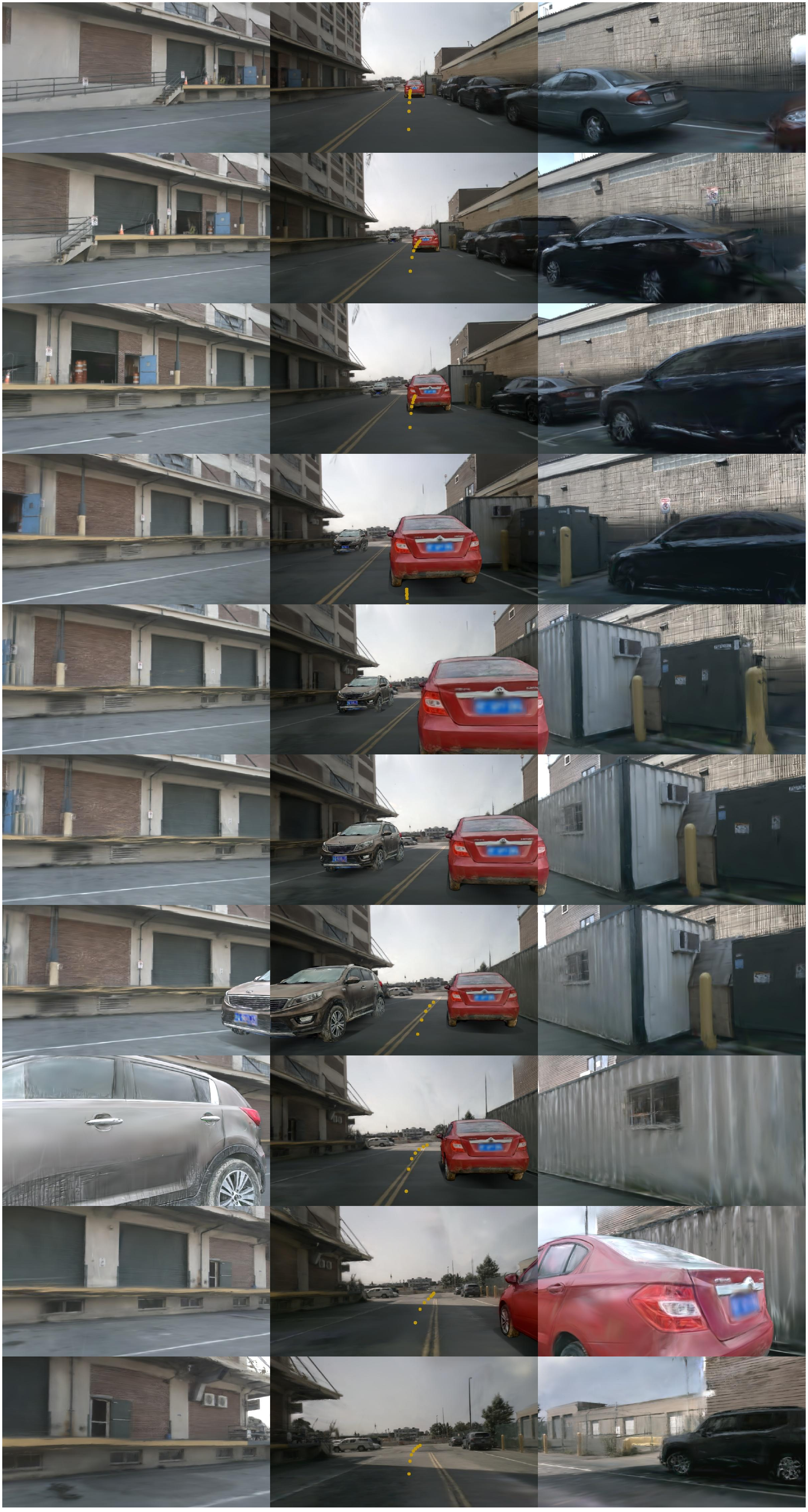}
        \caption{nuScenes~\citep{nuscenes}.}
        \label{app:fig:hugsim_nuscenes}
    \end{subfigure}
    \hfill
    \begin{subfigure}[t]{0.44\textwidth}
        \centering
        \includegraphics[width=\linewidth,trim=180 190 135 165,clip]
        {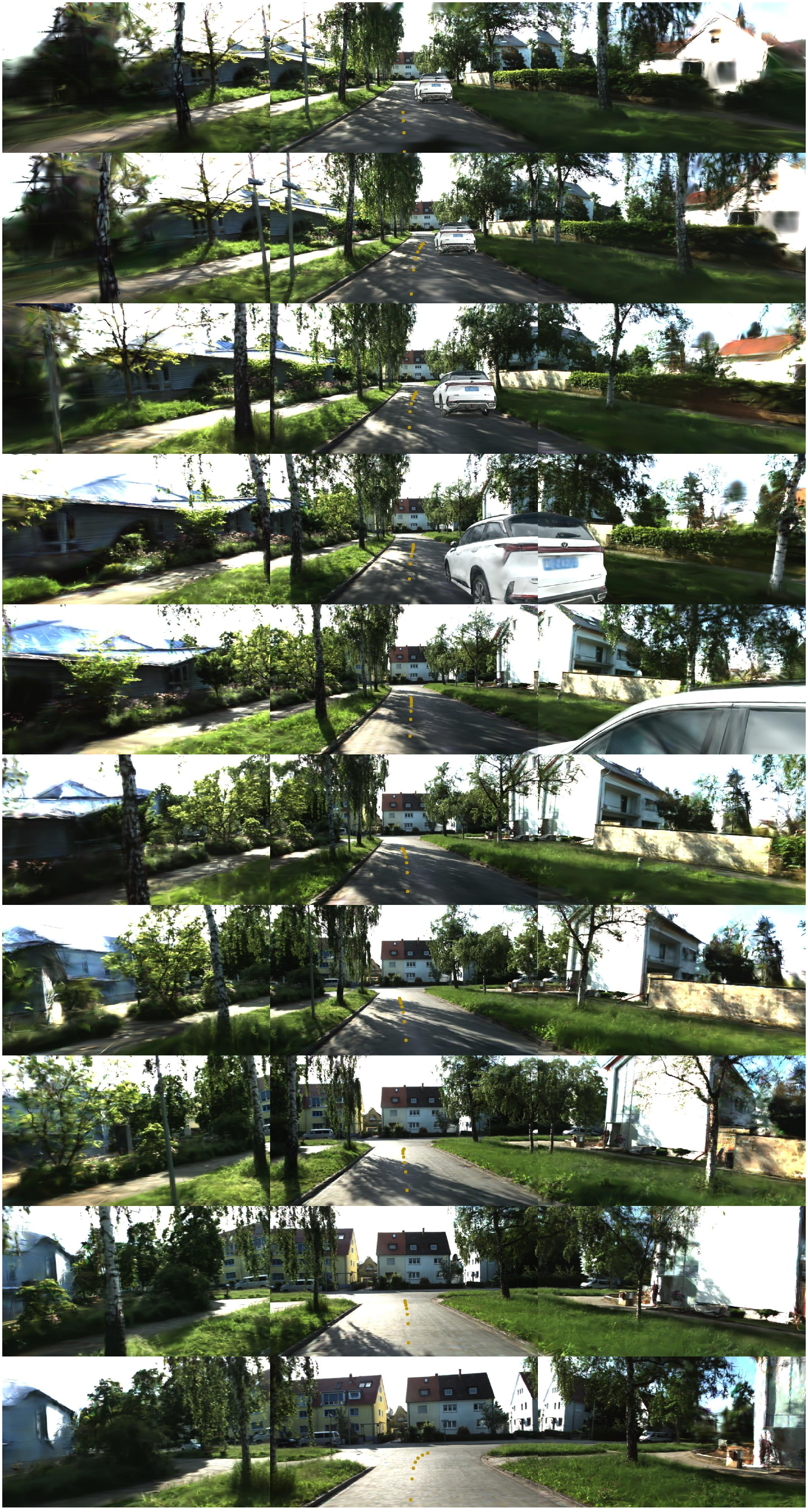}
        \caption{KITTI-360~\citep{kitti360}.}
        \label{app:fig:hugsim_kitti360}
    \end{subfigure}
    \caption{HUGSIM trajectory-planning visualizations on nuScenes and
    KITTI-360. Yellow trajectory points denote the trajectories planned
    by our model.}
    \label{app:fig:hugsim_nuscenes_kitti360}
\end{figure}

\clearpage
\begin{figure}[H]
    \centering
    \begin{subfigure}[t]{0.48\textwidth}
        \centering
        \includegraphics[width=\linewidth,trim=180 190 135 165,clip]
        {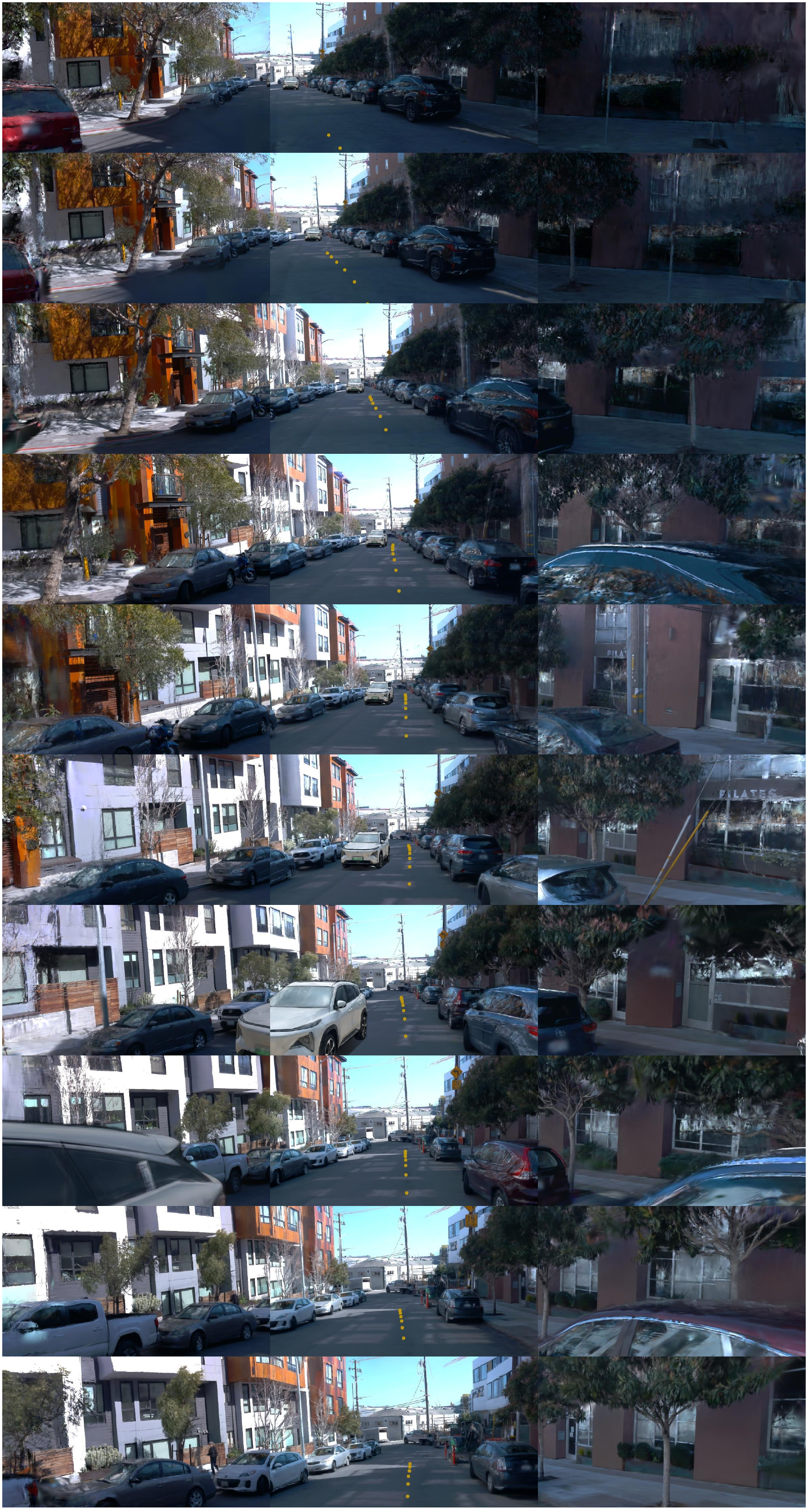}
        \caption{Waymo~\citep{waymo}.}
        \label{app:fig:hugsim_waymo}
    \end{subfigure}
    \hfill
    \begin{subfigure}[t]{0.48\textwidth}
        \centering
        \includegraphics[width=\linewidth,trim=180 190 135 165,clip]
        {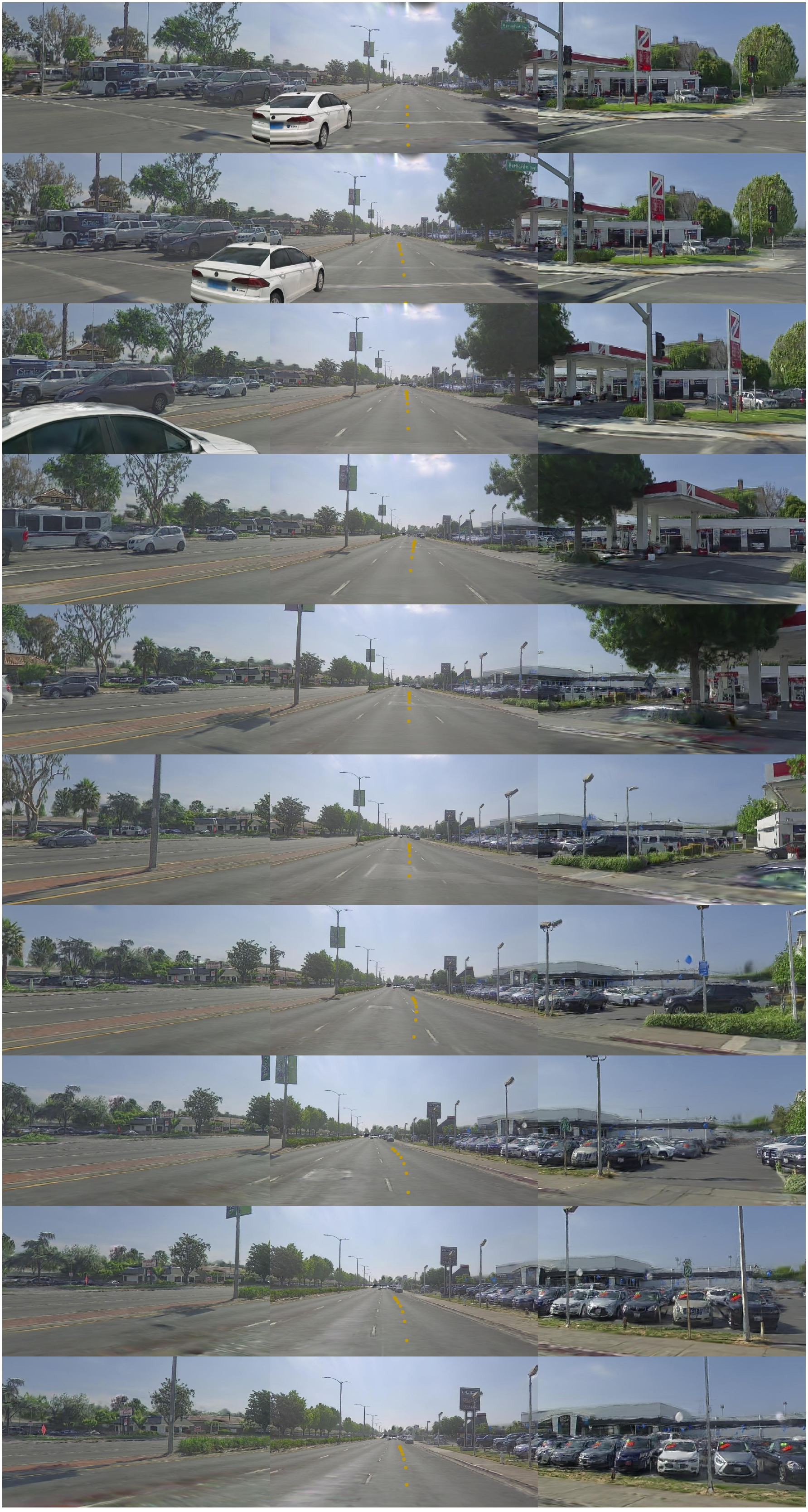}
        \caption{PandaSet~\citep{pandaset}.}
        \label{app:fig:hugsim_pandaset}
    \end{subfigure}
    \caption{HUGSIM trajectory-planning visualizations on Waymo and
    PandaSet. Yellow trajectory points denote the trajectories planned
    by our model.}
    \label{app:fig:hugsim_waymo_pandaset}
\end{figure}

\clearpage
\begingroup
\small
\renewcommand{\refname}{References for Supplementary Material}
\putbib[supplement]
\endgroup

\endgroup